\documentclass[journal]{IEEEtai}

\usepackage[colorlinks,urlcolor=blue,linkcolor=blue,citecolor=blue]{hyperref}
\usepackage{cite}
\usepackage{graphicx}
\usepackage{amsmath,amssymb,amsfonts}
\usepackage{algorithm}
\usepackage{algpseudocode}
\usepackage{graphicx}
\usepackage{textcomp}
\usepackage{xcolor}
\usepackage{url}
\usepackage{array}
\usepackage{booktabs}
\usepackage{threeparttable}
\usepackage{float}

\makeatletter
\let\OldStatex\Statex
\renewcommand{\Statex}[1][3]{%
  \setlength\@tempdima{\algorithmicindent}%
  \OldStatex\hskip\dimexpr#1\@tempdima\relax}
\makeatother

\usepackage{multirow}
\usepackage{epstopdf}

\begin{document}
\title{InFIP: An Explainable DNN Intellectual Property Protection Method based on Intrinsic Features}

\author{Mingfu Xue, Xin Wang, Yinghao Wu, Shifeng Ni, Yushu Zhang, and Weiqiang Liu

\thanks{M. Xue, X. Wang, Y. Wu, S. Ni,  Y. Zhang are with the College of Computer Science and Technology, Nanjing University of Aeronautics and Astronautics, Nanjing, China.}
\thanks{W. Liu is with the College of Electronic and Information Engineering, Nanjing University of Aeronautics and Astronautics, Nanjing, China.}}

\markboth{}
{M. Xue \MakeLowercase{\textit{et al.}}: InFIP: An Explainable DNN Intellectual Property Protection Method based on Intrinsic Features}

\maketitle

\begin{abstract}
Intellectual property (IP) protection for Deep Neural Networks (DNNs) has raised serious concerns in recent years. Most existing works embed watermarks in the DNN model for IP protection, which need to modify the model and lack of interpretability.
In this paper, for the first time, we propose an interpretable intellectual property protection method for DNN based on explainable artificial intelligence.
Compared with existing works, the proposed method does not modify the DNN model, and the decision of the ownership verification is interpretable.
We extract the intrinsic features of the DNN model by using Deep Taylor Decomposition.
Since the intrinsic feature is composed of unique interpretation of the model's decision, the intrinsic feature can be regarded as fingerprint of the model.
If the fingerprint of a suspected model is the same as the original model, the suspected model is considered as a pirated model.
Experimental results demonstrate that the fingerprints can be successfully used to verify the ownership of the model and the test accuracy of the model is not affected.
Furthermore, the proposed method is robust to fine-tuning attack, pruning attack, watermark overwriting attack, and adaptive attack.

\end{abstract}

\begin{IEEEkeywords}
Intellectual Property, Deep Neural Network, Fingerprint, Deep Taylor Decomposition, Intrinsic Feature.
\end{IEEEkeywords}

\section{Introduction}
\IEEEPARstart{D}{eep} Neural Networks (DNN) have been widely deployed in many domains.
Training a high-performance DNN model requires a large amount of labeled data, computing resources and expert knowledge, and the training process is very time-consuming \cite{9645219}.
Hence, the well-trained DNN model can be considered as valuable intellectual property (IP) of the model owner and needs to be protected from being infringed.

Most existing DNN IP protection methods (e.g., \cite{uchida2017embedding, RouhaniCK19, zhang2018protecting, adi2018turning}) aim to embed the watermark into the DNN model in the training stage.
For example, parameter-based work \cite{uchida2017embedding} embed watermark into the parameters of the model.
Backdoor-based methods \cite{zhang2018protecting, adi2018turning} use backdoor instances with specific labels to verify the ownership of the DNN model.
However, all the existing methods lack of interpretability, which is emerged as an important requirement for DNN intellectual property protection.

In this paper, we propose the first interpretable DNN \underline{I}ntellectual \underline{P}roperty protection method based on \underline{In}trinsic \underline{F}eatures (InFIP), which does not require to modify the DNN model.
We extract intrinsic features of the DNN model as its fingerprints for ownership verification.
The intrinsic features of the DNN model are extracted by Deep Taylor Decomposition (DTD) \cite{montavon2017explaining} method.
In DTD, the neurons of the model can be decomposed on input variables (i.e., pixels).
The decomposition of the neurons are aggregated into a saliency map, which is regarded as the fingerprint of the model.
Since the fingerprint indicates the relevance between each pixel in the image and the decision of the model,
the fingerprint can uniquely represent the DNN model, which is used to verify the ownership of the model.
The structural similarity index measure (SSIM) \cite{wang2004image} is used to calculate the similarity between the fingerprints of the original model and the suspected model.
If the average SSIM is greater than the threshold $T$, the model owner can claim the ownership of the suspected model.
Otherwise, the suspected model is not the pirated model.
Experimental results demonstrate that the proposed method can effectively verify the ownership of the model.
Moreover, when the model is modified by fine-tuning attack, pruning attack, watermark overwriting attack or adaptive attack, the fingerprint is still robust for ownership verification.

The main contributions of this paper are fourfold:
\begin{itemize}
\item{For the first time, we propose an explainable intellectual property protection method for DNN.
Specifically, we utilize DTD to extract the intrinsic features of the model.
Then the intrinsic features is converted as fingerprint images which can uniquely represent the DNN model.
These fingerprints are used to verify the ownership of the model.}
\item{Compared with most existing DNN IP protection methods, we extract fingerprints of the DNN model for ownership verification instead of embedding any watermarks into the DNN model.
In other words, the fingerprint is extracted from the clean DNN model, thus, the proposed IP protection method does not require to modify the DNN model.}
\item{The computational overhead of the proposed method is much lower than existing watermarking methods.
Specifically, existing watermarking methods require to fine-tune the model or even train the model from scratch to embed the watermark.
In comparison, the proposed method extracts fingerprints from models by inputting a batch of images, which will not introduce any extra training.}
\item{The proposed method is demonstrated to be robust against model fine-tuning attack, model pruning attack, watermark overwriting attack and adaptive attack.
  }
\end{itemize}

The rest of this paper is organized as follows.
Related works on DNN IP protection are reviewed in Section \ref{related work}.
The proposed method is presented in Section \ref{proposed method}.
Experimental results are discussed in Section \ref{experiments}.
This paper is concluded in Section \ref{conclusion}.

\section{Related Works}\label{related work}
Existing works on DNN intellectual property protection can be divided into three mechanisms: parameter-based method, backdoor-based method and fingerprint-based method \cite{9645219}.

\textit{(i) Parameter-based Method.} Uchida \textit{et al.} \cite{uchida2017embedding} use parameter regularizer to embed watermark in the convolutional layer of DNN model.
They extract the watermark from the model for DNN IP verification.
Rouhani \textit{et al.} \cite{RouhaniCK19} embed watermark into the Probability Distribution Function (PDF) of the activations of the model by fine-tuning the model on a set of specific data.
During inference stage, when the set of specific data is input into the model, the statistical mean value of the activations of the model is calculated for ownership verification.
Kuribayashi \textit{el al.} \cite{KuribayashiTF20} embed watermark into the selected weights in the fully-connected layer.
Then they use quantifiable method (i.e., DM-QIM) to replace original weights with watermarked weights.
Finally, they fine-tune the model to reduce the influence of watermark on test accuracy of the model.
However, for the parameter-based method, when the length of embedded watermark is long, the test accuracy of the protected model on clean images will be slightly degraded  \cite{CaoJG21}.
Furthermore, since the parameter-based watermark will change the distribution of the weights of the model, this method is vulnerable to statistical analysis attack\cite{WangK19}.

\textit{(ii) Backdoor-based Method.} Zhang \textit{et al.} \cite{zhang2018protecting} propose backdoor-based method to protect the IP of the DNN model.
In the training stage, they train the DNN model with backdoor instances to embed the backdoor.
In the inference stage, when the backdoor instances are input into the DNN model, the DNN model will output specified prediction so as to verify the ownership.
Adi \textit{et al} \cite{adi2018turning} propose a black-box watermark method based on backdoor.
When the model is suspected of being pirated, the model owner inputs the query set into the suspected model without access to the internal parameters of the model.
If the suspected model predicts most backdoor instances as the specified label, the suspected model is considered as a pirated model.
Xue \textit{et al.} \cite{xue2022active} assign an additional class to watermark key instances, where the instances are embedded with users' identity information via image steganography for identity authentication.
If the watermark accuracy is greater than the predefined threshold, the model owner can claim that the suspected model is the pirated model.
Otherwise, the suspected model is not the pirated model.
Besides, the user's identity can be verified by extracting fingerprint from the watermark images through image steganography method.
However, the works \cite{ShafieinejadLWL21,Chen0BDJLS21,AikenKWR21} indicate that the backdoor-based watermark \cite{zhang2018protecting,adi2018turning,xue2022active} may be removed by the fine-tuning attack or pruning attack.

\textit{(iii) Fingerprint-based Method.}
Merrer \textit{et al.}\cite{MerrerPT20} use adversarial example as watermark for DNN IP protection.
They slightly alter the decision boundary of the model to find some data points, which are used to remotely query the suspected model for DNN ownership verification.
If the accuracy of the adversarial example set is greater than the threshold, the suspected model is considered as the pirated model.
Cao \textit{et al.} \cite{CaoJG21} select data points near the decision boundary and use these data points as fingerprint to protect the IP of the DNN model.
If the suspected model classifies most data points as wrong labels, it is considered as a pirated model.
Pan \textit{et al.}\cite{PanYZY22} use prediction vectors of protected model and other models on \textit{adaptive fingerprints} (a set of selected images) to train an extra binary classifier.
By inputting the prediction vectors of suspected model on \textit{adaptive fingerprints} into the extra binary classifier, the binary classifier can be used to distinguish the ownership of the suspected model.

However, all the existing works, e.g., \cite{uchida2017embedding,KuribayashiTF20, RouhaniCK19,zhang2018protecting, adi2018turning, xue2022active, CaoJG21, MerrerPT20, PanYZY22}, lack of interpretability.
Moreover, most works \cite{uchida2017embedding,KuribayashiTF20, RouhaniCK19, zhang2018protecting,adi2018turning,xue2022active,MerrerPT20} require to modify the DNN model to implement their DNN IP protection methods.
As a comparision, we propose an interpretable DNN IP protection method which does not require to modify the DNN model.
The intrinsic features of model are extracted as the model's fingerprints through an eXplainable Artificial Intelligence (XAI) method, DTD \cite{montavon2017explaining}, which makes the extracted fingerprints can uniquely represent the specific model.

\section{Proposed Method}\label{proposed method}
\subsection{Problem Formulation}
\begin{figure}[!htbp]
  \centering
  \includegraphics[width=.48\textwidth]{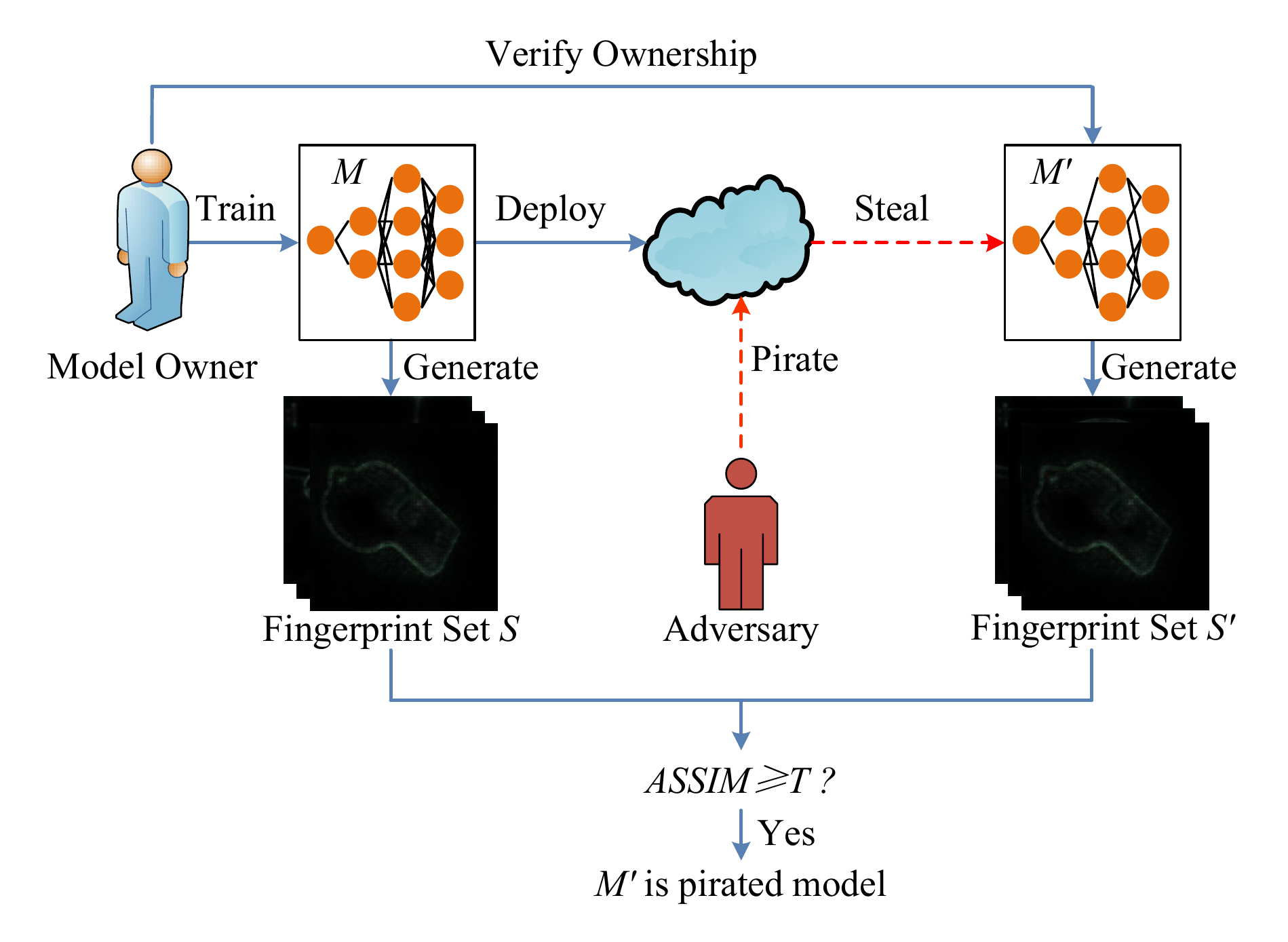}\\
  \caption{The application scenario of the proposed method.}
  \label{problem_formulation}
\end{figure}

As shown in Fig. \ref{problem_formulation}, the model owner trains the DNN model and deploys the well-designed model $M$ as a cloud service.
Then the model owner uses private key instance set (a set of images) to generate the fingerprint set $S$ of the model $M$ and saves $S$ privately.
However, the adversary may steal the model and then illegally distribute or resell the model to others for profits.
When the model owner wants to verify the ownership of the suspected model, he can generate the fingerprint set $S'$ of ${M'}$.
The fingerprint set $S'$ is used to verify whether ${M'}$ is derived from ${M}$.
The model owner uses $SSIM$ \cite{wang2004image} to compare the similarity between $S$ and $S'$, and claim ownership of ${M'}$ according to the average value of $SSIM$.

\subsection{Overall Flow}
\begin{figure}[!htbp]
\centering
\includegraphics[width=0.5\textwidth]{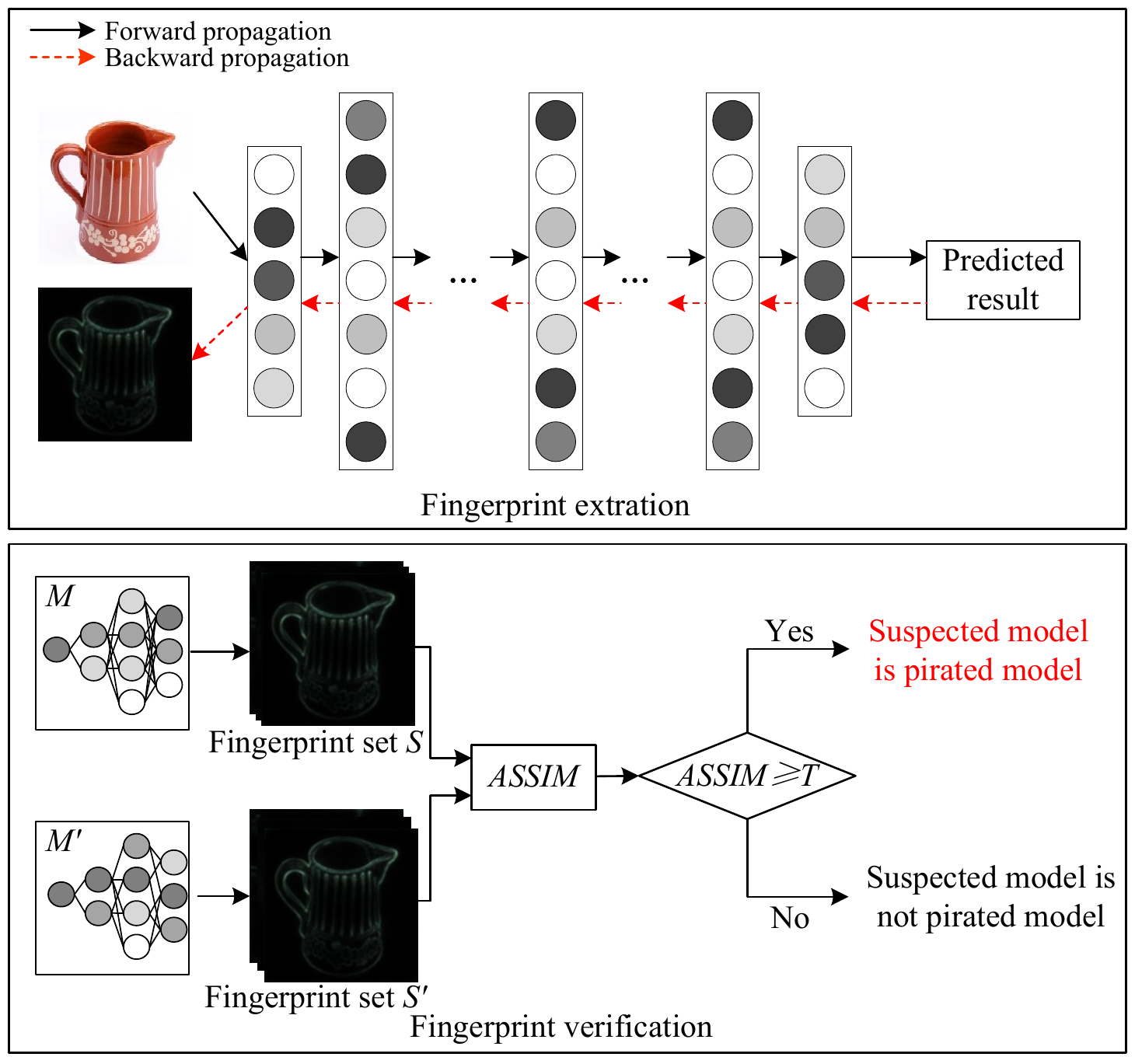}\\
\caption{Overview of the proposed DNN intellectual property protection method}
\label{overview}
\end{figure}

In this section, the proposed DNN intellectual property method is elaborated.
As shown in Fig. \ref{overview}, the proposed IP protection method consists of three steps: (i) original model's fingerprint extraction; (ii) suspected model's fingerprint extraction; (iii) DNN IP verification.
Algorithm \ref{alg:FE} summarizes the whole process of the proposed method.
As show in Algorithm \ref{alg:FE}, in the first step, the fingerprint is extracted from the original model.
We randomly select $N$ images to generate key instance set $K$.
When the set $K$ is input into the model, the DTD \cite{montavon2017explaining} is exploited to generate fingerprint set $S$ of the original model ${M}$.
In the second step, the fingerprint is extracted from the suspected model.
When the model is suspected to be a pirated model, the fingerprint set $S'$ of the model ${M'}$ is generated via DTD.
In the third step, the fingerprints are used to verify the ownership of the model ${M'}$.
The similarity between $S$ and $S'$ is calculated by $SSIM$.
If the average $SSIM$ between $S$ and $S'$ is greater than the pre-defined threshold $T$, the ${M'}$ is considered as a pirated version of ${M}$.
Otherwise, the model ${M'}$ is not the pirated version of ${M}$.
In the following sections, we will elaborate the process of fingerprint extraction and fingerprint verification respectively.

\begin{algorithm}
\caption{The Proposed InFIP Method.}
\label{alg:FE}
\begin{algorithmic}[1]
\Require
    \Statex[0] (1) key instances set ${K}=\{ {k_1},{k_2}, \ldots ,{k_N}\} $;
    \Statex[0] (2) original model ${M}$;
    \Statex[0] (3) suspected model $M^{\prime}$;
    \Statex[0] (4) hyper-parameter $\lambda$;
    \Statex[0] (5) threshold $T$;
\Ensure whether $M^{\prime}$ is a pirated model.
\State \textbf{Initialize} $temp \leftarrow 0$;
\State predictions of $M$ on $K$ is $R=\{{r}_1,{r}_2,\ldots, {r}_N\}$
\State predictions of $M^{\prime}$ on $K$ is $R^{\prime}=\{r^{\prime}_1,{r^{\prime}_2},\ldots, {r^{\prime}_N}\}$
\For{$i=1 \to N$}
\State $ InF(k_i) \gets DTD({M}, k_i, r_i)$
\State $s_i \gets Image(InF(k_i), \lambda)$
\EndFor
\For{$i=1 \to N$}
\State $ InF(k_i) \gets DTD(M^{\prime}, k_i, r^{\prime}_i)$
\State $s^{\prime}_i \gets Image(InF(k_i), \lambda)$
\EndFor
\For{$i=1 \to N$}
\State $temp \gets temp + SSIM(s_i, s^{\prime}_i)$
\EndFor
\State $ASSIM = \frac{1}{N}\times temp$\
\If{$ASSIM \ge T$}
{\\
   \Return $M^{\prime}$ is a pirated model\;
}
\Else {\\
    \Return $M^{\prime}$ is not a pirated model\;
}
\EndIf
\end{algorithmic}
\end{algorithm}

\subsection{Fingerprint Extraction}\label{dtd}
Before verifying the ownership of the model, the fingerprint of original model requires to be extracted and saved for comparison later.
The detailed process of extracting fingerprint is elaborated as follows.

Deep Taylor Decomposition (DTD) \cite{montavon2017explaining} is usually used to explain the decision of model through pixel-wise decomposition.
In this paper, we propose to utilize DTD to extract the intrinsic features as the fingerprints of the model.
Specifically, for an input image $x$, the predicted result $g(x)$ is propagated backwards from the output layer to the input layer, where $g(\cdot)$ is a differentiable function.
$g(x)$ can be decomposed at the root point $x'$ and $ g(x')=0$.
The DTD can be formulated as \cite{montavon2017explaining}:
\begin{equation}\label{DTD}
g(x) = g(x') + \sum\limits_p {g'({x_p})}  + \varepsilon  \approx \sum\limits_p {g'({x_p})}
\end{equation}
where $\varepsilon$ denotes the 2-th and higher order derivatives of $g(x_p)$, which can be omitted in the following deduction.
$p$ denotes the $p$-th pixel of the image $x$, ${g'(x_p)}$ denotes the relevance score between the $p$-th pixel and predicted result.
${g'(x_p)}$ can be formulated as follows\cite{montavon2017explaining}:

\begin{equation}\label{INFP}
{g'}(x_p) = \frac{{\partial g}}{{\partial {x_p}}}{|_{x = x'}} \cdot ({x_p} - {{x'}_p})
\end{equation}

The intrinsic feature matrix $InF(x)$ of the model is formalized as follows:
\begin{equation}\label{INF}
{InF(x)} = DTD(M,x,r)
\end{equation}
where $x$ denotes the key instance, $r$ denotes the corresponding predicted result of $M$.
Then the matrix $InF(x)$ is converted as fingerprint $s$, which can be formalized as follows:
\begin{equation}\label{image}
{s} =Image(InF(x), \lambda)
\end{equation}
where the value of $\lambda$ controls the magnification, which will be discussed in Section \ref{hyper}.
In practice, we use a set of clean images as the key instances set $K$ to extract the fingerprints from the model, where $K$ consists of the randomly selected $N$ images.
Specifically, the key instance set $K$ is input into the model $M$ to generate the corresponding predicted result set $R$.
Then, the set $R$ is back-propagated to generate the fingerprint set $S$ of the model via DTD, and the generated fingerprint set $S$ will be used for comparison with that of the suspected model.
For a suspected model ${M'}$, the same method is used to generate the fingerprint set $S'$.
The fingerprint set $S'$ of suspected model $M^{\prime}$ will be compared with fingerprint set $S$ of the original model $M$.

\subsection{Fingerprint Verification}
When the image is input into the same model, the fingerprints of the model generated by DTD are consistent.
When the image is input into the fine-tuned and pruned models, the fingerprints of the models generated by DTD are highly similar.
Thus, the fingerprint can uniquely represent the model.
We use the structural similarity index measure (SSIM) \cite{wang2004image} to calculate the similarity between $S$ of the protected model and $S'$ of the suspected model.
The pre-defined threshold $T$ is used to determine whether the model ${M'}$ is the pirated version of the model ${M}$.

$SSIM$ \cite{wang2004image} evaluates the similarity of the images in terms of luminance, contrast and structure.
$SSIM$ is calculated as follows \cite{wang2004image}:
\begin{equation}\label{equation_5}
SSIM(x,y) = \frac{{(2{\phi _x}{\phi _y} + {c_1})(2{\varphi _{xy}} + {c_2})}}{{(\phi _x^2 + \phi _y^2 + {c_1})(\varphi _x^2 + \varphi _y^2 + {c_2})}}
\end{equation}
where ${\phi _x}$ and ${\phi _y}$ denote the mean of pixel values in the image $x$ and the image $y$ respectively, $\varphi _{xy}$ denotes the covariance between the input $x$ and the input $y$, $\varphi_{x}$ and $\varphi_{y}$ denote the standard deviation of images $x$ and $y$ respectively. ${c_1}$ and ${c_2}$ are empirically set to 0.0001 and 0.0009 respectively.

When the model owner suspects that the model ${M'}$ is a pirated version of the model ${M}$, he can use $SSIM$ to calculate the similarity between fingerprint set $S$ of the model ${M}$ and fingerprint set $S'$ of the model ${M'}$.
If $SSIM(S, S')\ge T$, the suspected model ${M'}$ is considered as a pirated version of the model ${M}$.
Otherwise, the suspected model ${M'}$ is not the pirated version of the model ${M}$.

\subsection{Why the Proposed Method Can Be Used to Verify the Ownership of the Model}
Deep Taylor Decomposition (DTD) \cite{montavon2017explaining} aims to explain the predictions of model by decomposition.
In other words, the decision of neural network is explained by calculating the relevance scores of the neurons from model's output layer to the model's input layer.
In this paper, we use DTD to extract the fingerprint of the model.
For a deep neural network, the activations of these neurons are unique, which can be used to uniquely represent the model.
By inputting a query image to activate these neurons, we can obtain the relevance scores that represent the activations of the model.
As a result, the fingerprint visualized from these relevance scores can uniquely represent the model.

\begin{figure}[!htbp]
  \centering
  \includegraphics[width=0.48\textwidth]{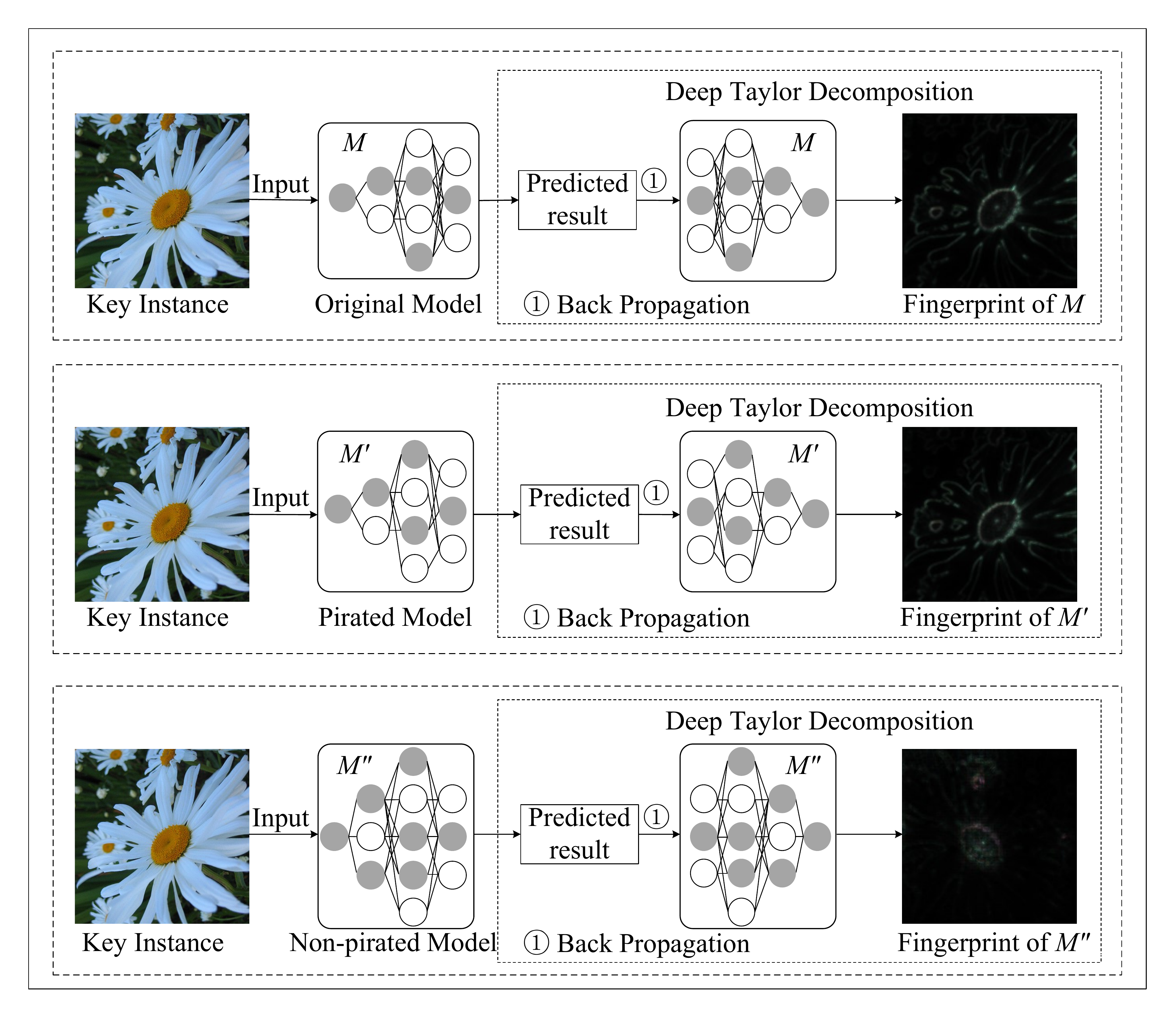}\\
  \caption{Examples of fingerprints generated from the original model $M$, pirated model $M^{\prime}$ and non-pirated model $M^{\prime\prime}$.}
  \label{fp_extra}
\end{figure}

Fig. \ref{fp_extra} shows the examples of the fingerprint of
model $M$, $M'$ and $M''$ respectively,
where $M$ denotes the protected model, $M'$ denotes the pirated model, $M''$ denotes another model that is not a pirated version of the model $M$.
Note that, the key instance set contains a set of images.
In Fig. \ref{fp_extra}, we only use one key instance to show the difference among the fingerprints generated from the model $M$, $M'$ and $M''$.
As shown in Fig. \ref{fp_extra}, the fingerprints of $M$ and $M'$ are indistinguishable, while the fingerprints of $M$ and $M''$ are clearly distinguishable.
It is demonstrated that the extracted fingerprint can uniquely represent the model and can be used to verify the ownership of the model.

\section{Experiments}\label{experiments}
\subsection{Experimental Setup}\label{setup}
In the experiments, CIFAR-10 \cite{krizhevsky2009learning1} and ImageNet \cite{DengDSLL009} datasets are used to evaluate the effectiveness of the proposed DNN IP protection method.
CIFAR-10 \cite{krizhevsky2009learning1} dataset consists of 10 classes.
Each class has 5,000 training images and 1,000 test images.
The size of each image is $32 \times 32$.
In ImageNet \cite{DengDSLL009} dataset, the size of each image is $224 \times 224$.
In the experiments, 100 classes randomly selected from the ImageNet dataset are used for model training and testing.

The models used in this experiments are VGG-16 model \cite{SimonyanZ14a} and ResNet-18 model \cite{he2016deep}.
Each model is trained on CIFAR-10 \cite{krizhevsky2009learning1} and ImageNet \cite{DengDSLL009} datasets respectively.
For VGG-16 model \cite{SimonyanZ14a}, the test accuracy (on clean images) of the model trained on CIFAR-10 and ImageNet datasets is 86.10\% and 73.82\%, respectively.
For ResNet-18 model \cite{he2016deep}, the test accuracy (on clean images) of the model trained on CIFAR-10 and ImageNet datasets is 82.02\% and 75.94\%, respectively.

Two evaluation metrics, test accuracy (TA) \cite{goodfellow2016deep} and average SSIM \cite{wang2004image} ($ASSIM$), are used to evaluate the performance of the proposed DNN IP protection method.
Given a test set, TA denotes the percentage of the images that are classified as correct labels among all test images \cite{goodfellow2016deep}.
$ASSIM$ denotes the average value of the SSIM between $S$ and $S'$.
It can be calculated by $ASSIM = \frac{1}{N}\sum\nolimits_{i = 1}^N {SSIM({s_i},{{s'}_i})} $, where $N$ denotes the number of fingerprints in the fingerprint set, $s_i$ and $s_i^{\prime}$ denote the $i$-th fingerprint in fingerprint set $S$ and $S^{\prime}$ respectively.
The pre-defined threshold $T$ is empirically set to be 0.85 for ownership verification.

\subsection{Experimental Results}\label{effectiveness}
\begin{figure}[!htbp]
\centering
\includegraphics[width=0.48\textwidth]{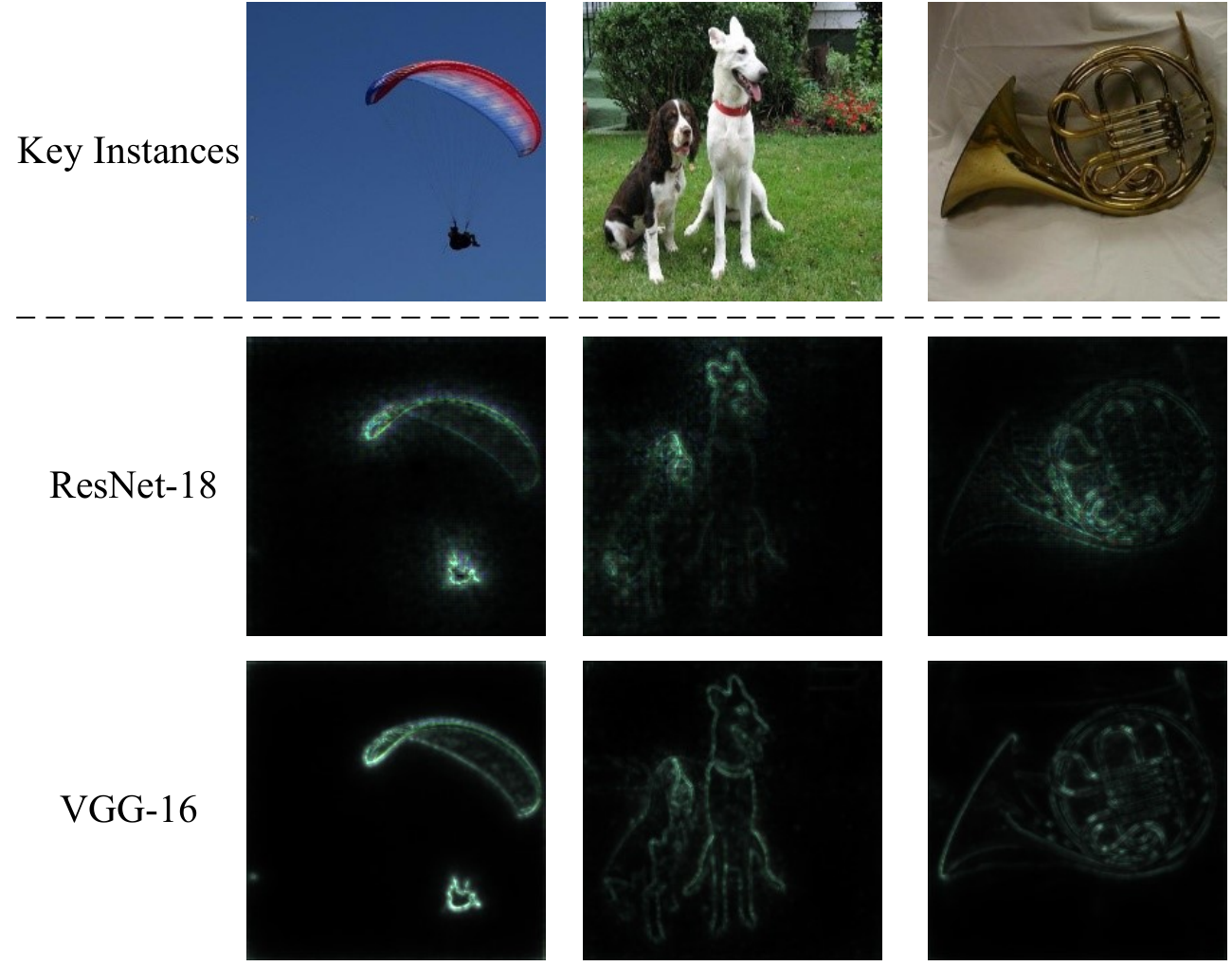}
\caption{Examples of fingerprints generated from ResNet-18 model and VGG-16 model.}
\label{fig_effectiveness}
\end{figure}

In this section, we evaluate the effectiveness of the proposed DNN IP protection method on CIFAR-10 \cite{krizhevsky2009learning1} and ImageNet \cite{DengDSLL009} datasets respectively.
The key instance set $K$ consists of 400 images, which are randomly selected from the test set of the dataset (CIFAR-10 or ImageNet).
The key instance set $K$ is input into the model for generating the corresponding fingerprint set.
On CIFAR-10 and ImageNet datasets, the fingerprint set is extracted from VGG-16 \cite{SimonyanZ14a} model and ResNet-18 \cite{he2016deep} model respectively.
The experimental results on CIFAR-10 and ImageNet datasets are shown in Table \ref{tab:effectiveness}.
When the fingerprint sets $S$ and $S'$ are generated from the same model (VGG-16 model or ResNet-18 model), the fingerprint is consistent.
Hence, the fingerprint can successfully demonstrate the ownership of the model.
The examples of fingerprints of the model on ImageNet dataset are shown in Fig. \ref{fig_effectiveness}.
The first row is the key instances.
The second row is the fingerprints that are extracted from the ResNet-18 model.
The third row is the fingerprints that are extracted from the VGG-16 model.
The $ASSIM$ between the fingerprint set generated from ResNet-18 model and that generated from VGG-16 model is 0.7310.
It is shown that the fingerprints generated from different models are totally different, which means the proposed method can successfully distinguish the original model from other models.
The reason is that, when the key instances set $K$ is input into different DNN models, the activated neurons are different.
Thus, the generated fingerprints are different.

In summary, the fingerprints extracted from the same model remain consistent and the fingerprints extracted from different models are totally different.
This demonstrates that the fingerprint extracted from the protected model can uniquely represent the model.
As a result, the proposed DNN IP protection method can effectively verify the ownership of the model.

\begin{table}[!htbp]
\newcommand{\tabincell}[2]{\begin{tabular}{@{}#1@{}}#2\end{tabular}}
\centering
\caption{Test Accuracy and \textit{ASSIM} on CIFAR-10 and ImageNet Datasets.}
\begin{tabular}{cccc}
\toprule
\textbf{Dataset} & \textbf{Model} & \textbf{TA} & \textit{ASSIM} \\
\midrule
\multirow{2}*{CIFAR-10}
& VGG-16 & 86.10\% & 1.0000\\
& ResNet-18 & 82.02\% & 1.0000 \\
\toprule
\multirow{2}*{ImageNet}
& VGG-16 & 73.82\% & 1.0000 \\
& ResNet-18 & 75.94\% & 1.0000 \\
\bottomrule
\end{tabular}
\label{tab:effectiveness}
\end{table}

\subsection{Parameters Discussion}\label{paradis}
In this section, we discuss the influence of different values of hyper-parameter $\lambda$ and the number of key instances on the performance of ownership verification.

\subsubsection{The Impact of Hyper-parameter \textit{$\lambda$}}\label{hyper}
We discuss the impact of the hyper-parameter $\lambda$, which is set to 1000, 5000, 7500, 10000, 12500 and 15000 respectively.
The model used in this experiment is ResNet-18.
The value of $\lambda$ controls the magnification from intrinsic features to the fingerprints, where the values of the intrinsic features are in the range of 0$\sim$1 and the values of the fingerprints are in the range of 0$\sim$255.
The key instance set $K$ is composed of 400 images, which is randomly selected from 100 classes of the test set of the ImageNet dataset and each class contains four images.
During the process of fingerprint extraction, $K$ and its predicted result set $R$ are used to generate the intrinsic features of the model ${M}$ by DTD.
The generated intrinsic features of the original model and the suspected model are visualized as fingerprints set $S$ and $S'$ respectively.

\begin{figure*}[!htbp]
\centering
\includegraphics[width=0.9\textwidth]{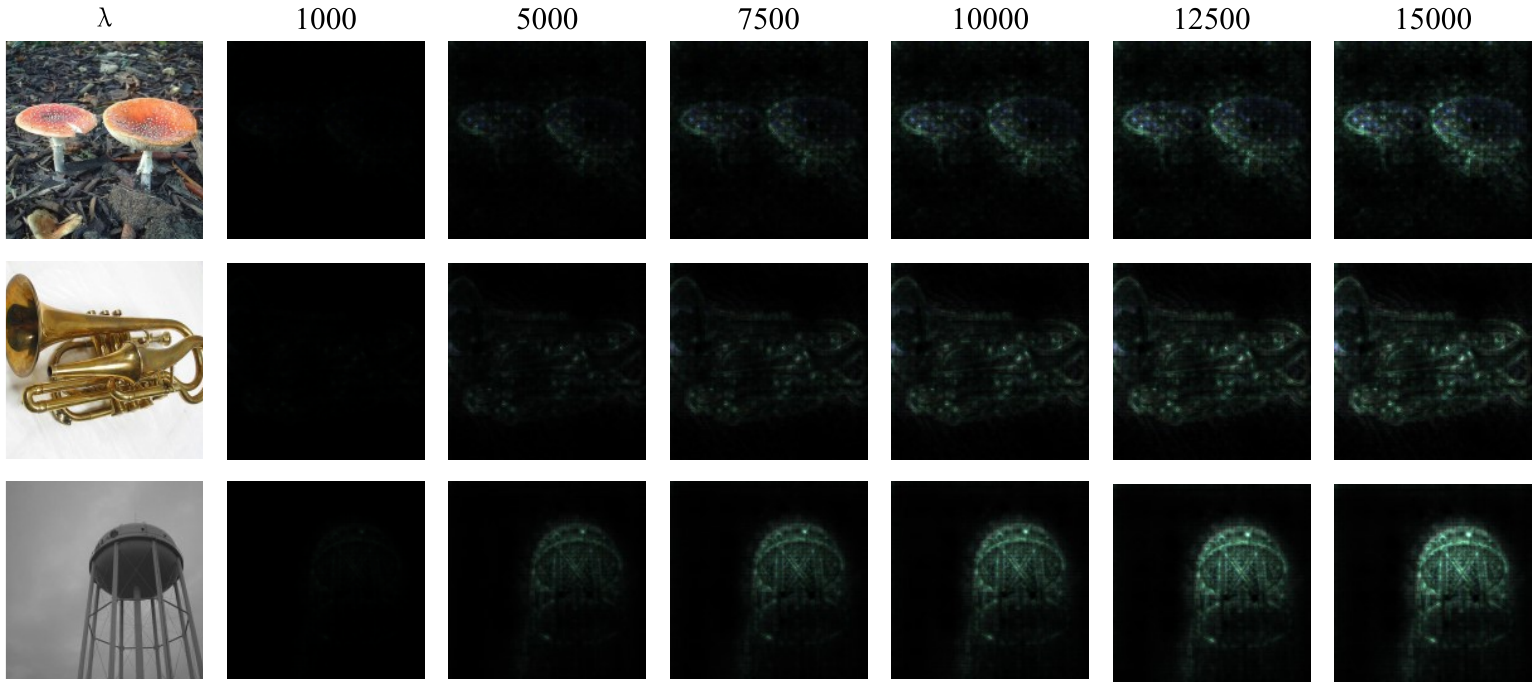}\\
\caption{Examples of fingerprints extracted from the ResNet-18 model with different values of $\lambda$ on ImageNet dataset.}
\label{fig:lambda}
\end{figure*}

As shown in Table \ref{tab:lambda}, without model modification attacks, as the value of $\lambda$ increases, the $ASSIM$ between $S$ and $S'$ is always 1.0000.
Thus, the model owner can verify that the suspected model is the pirated version regardless of the value of $\lambda$.
Under model modification attacks, with the increase of $\lambda$, the $ASSIM$ slightly degrades but is still much larger than the threshold $T$.
Specifically, under the fine-tuning attack, the $ASSIM$ is 0.9794 and 0.9718 when $\lambda$ is 15,000 and 1,000, respectively.
When the model is attacked by watermark overwriting, the $ASSIM$ changes from 0.9960 to 0.9816 as the hyper-parameter $\lambda$ increases.
The experimental results show that the proposed method can successfully verify the ownership of the model regardless of the values of $\lambda$, even though the model undergoes model modification attacks.
Furthermore, 5,000 is a suitable setting for hyper-parameter $\lambda$.
The reason is that the pixel values of the fingerprint increase as the value of $\lambda$ increases.
However, as shown in Fig. \ref{fig:lambda}, when the value of $\lambda$ is set to 1,000, the fingerprints are almost invisible.
When $\lambda$ is set to 5,000, all pixel values of fingerprints are less than 255 and the fingerprints are visible.

\begin{table*}[!htbp]
\centering
\caption{\textit{ASSIM} of the Fingerprints of the Original Model and Models Subjected to Model Modification Attacks under Different Hyper-parameter $\lambda$.}
\begin{tabular}{ccccccc}
\toprule
\multirow{2}[3]{*}{\textbf{Model Modification Attack} }
& \multicolumn{6}{c}{$\lambda$} \\
\cmidrule{2-7}
& {1000}&{5000}&{7500}&{10000}&{12500}&{15000} \\
\midrule
None&1.0000&1.0000&1.0000&1.0000&1.0000&1.0000 \\
Fine-tuning Attack & 0.9718 & 0.9877 & 0.9850 & 0.9828 & 0.9806 & 0.9794 \\
Pruning Attack & 0.9909 & 0.9966 & 0.9958 & 0.9983  & 0.9948  & 0.9944  \\
Watermark Overwriting Attack & 0.9960& 0.9887 & 0.9863  & 0.9845 & 0.9829 & 0.9816 \\
\bottomrule
\end{tabular}%
\label{tab:lambda}
\end{table*}

\subsubsection{The Number of Key Instances}\label{num}
In this section, we discuss the influence of the number of key instances on ownership verification.
We select 1, 4, 8, 12, 16 and 20 images from each class, i.e., we choose 100, 400, 800, 1,200, 1,600 and 2,000 images from the test set of the ImageNet as the key instance set $K$.
The experiment is performed on ResNet-18 model.
We use the set $K$ and its corresponding predicted result set $R$ to generate the intrinsic features of the model by DTD.
The generated intrinsic features of the original model and the suspected model are visualized as fingerprints set $S$ and $S'$ respectively, where $\lambda$ is set to 5000.

As shown in Table \ref{tab:number}, the $ASSIM$ increases as the number of key instances increases from 100 to 400.
The $ASSIM$ is almost unchanged when the number of key instances increases from 400 to 2000.
It is demonstrated that, the ownership of model can be successfully verified by using any number of instances.
When the number of instances is higher than 400, the $ASSIM$ reaches a stable high value (around 0.9870).
Therefore, 400 is a suitable setting for verifying the ownership of the DNN model.

\begin{table*}[!htbp]
\centering
\caption{\textit{ASSIM} of Fingerprints of the Original Model and Models Subjected to Model Modification Attacks under Different Numbers of Key Instances.}
\begin{tabular}{ccccccc}
\toprule
\multirow{2}[3]{*}{\textbf{Model Modification Attack }} & \multicolumn{6}{c}{\textbf{The Number of Key Instances}} \\
\cmidrule{2-7}
 &{100} & {400} & {800} & {1200} & {1600} & {2000} \\
\midrule
None&1.0000 & 1.0000 & 1.0000 & 1.0000 & 1.0000 & 1.0000 \\
Fine-tuning Attack&0.9027 & 0.9877 & 0.9875 & 0.9875 & 0.9874 & 0.9873 \\
Pruning Attack&0.9650 & 0.9966 & 0.9966 & 0.9967 & 0.9967 & 0.9966 \\
Watermark Overwriting Attack&0.9865 & 0.9887 & 0.9885 & 0.9886 &0.9886 & 0.9886 \\
\bottomrule
\end{tabular}%
\label{tab:number}
\end{table*}

\subsection{Robustness}\label{robustness}

Since the model modification attacks can change the parameters of the DNN and may affect the intrinsic features of the DNN model,
the extracted fingerprints may be slightly changed.
In this section, we evaluate the robustness of the proposed DNN IP protection method against fine-tuning attack \cite{PittarasMMP17}, pruning attack \cite{HanPTD15}, watermark overwriting attack \cite{RouhaniCK19} and adaptive attack.
We perform the experiments on ImageNet and CIFAR-10 datasets respectively.
There are 400 images in key instance set $K$.
We extract the fingerprints from VGG-16 and ResNet-18 models and their corresponding pirated versions respectively.

\subsubsection{Fine-tuning Attack}\label{fine_tune}
\begin{figure}[!htbp]
  \centering
    \includegraphics[width=.49\textwidth]{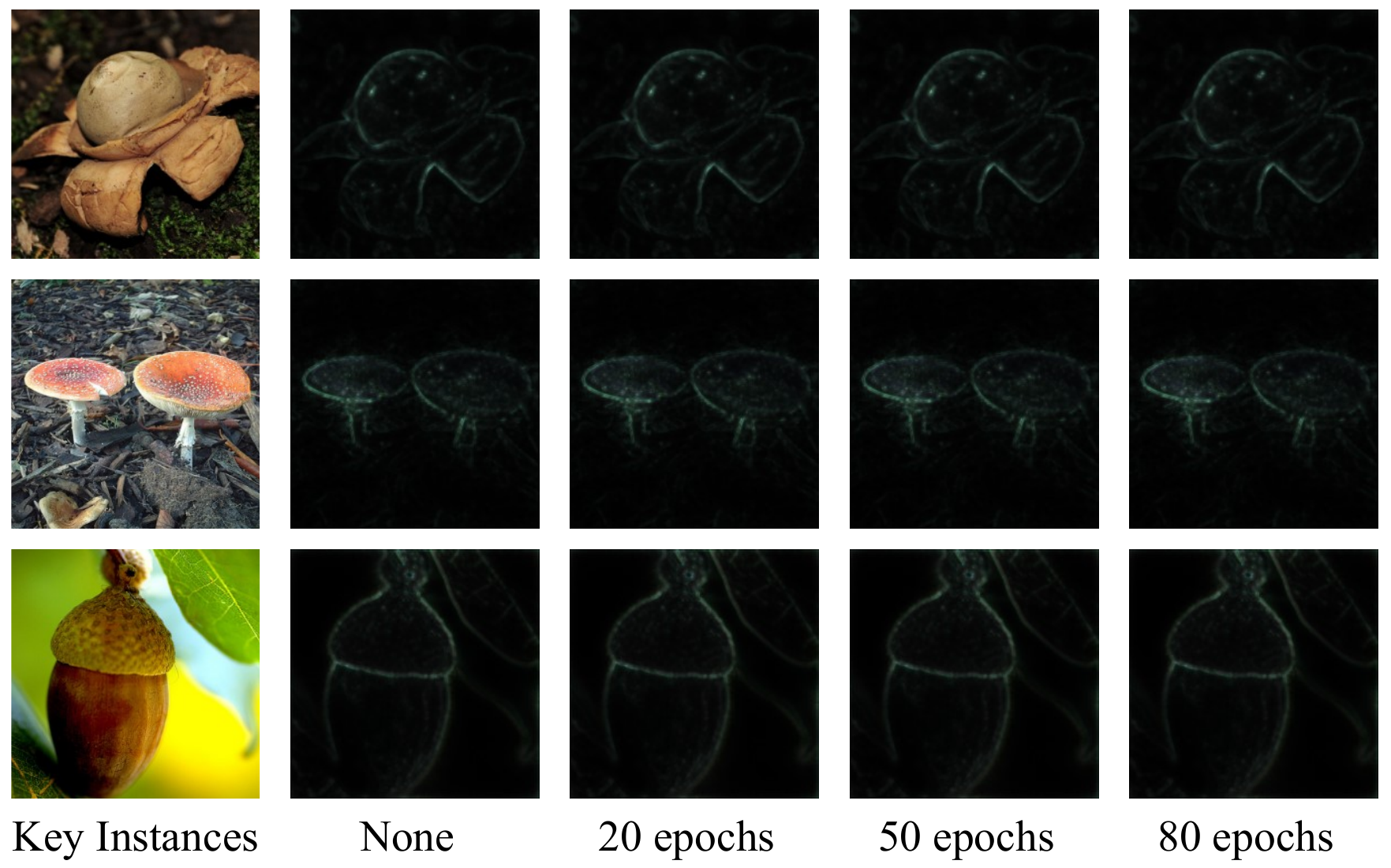}\\
  \caption{Examples of fingerprints of the models after different epochs of fine-tuning attack}
  \label{fig_finetune}
\end{figure}

Fine-tuning attack \cite{PittarasMMP17} can modify the model with a few amount of training data.
The attacker may try to remove the watermark in the DNN model via fine-tuning the model.
For fine-tuning attack, 2,000 images are randomly selected from the test set of ImageNet dataset to fine-tune the model.
We fine-tune VGG-16 model and ResNet-18 model for 20, 50 and 80 epoches respectively.
The robustness of the proposed DNN IP protection method against fine-tuning attack is shown in Table \ref{tab:fine-tune}.
Under fine-tuning attack, the parameters of model will be slightly changed.
Hence, the extracted fingerprint is slightly changed as well.
The $ASSIM$ of $S$ and $S'$ is slightly lower than 1.0000 but much greater than threshold $T$, which indicates that the fine-tuned model is the pirated model.
Specifically, when the VGG-16 model is fine-tuned for 20, 50 and 80 epochs respectively, the $ASSIM$ is 0.9992, 0.9989, 0.9987 respectively.
When the ResNet-18 model is fine-tuned for 20, 50 and 80 epochs, the $ASSIM$ is 0.9366, 0.9064, 0.8964 respectively.
It is demonstrated that, although the parameters of the model have been changed by the fine-tuning attack, the extracted fingerprints of fine-tuned model is still consistent to that of the original model.
Fig. \ref{fig_finetune} shows some examples of fingerprints of the models after different epochs of fine-tuning attack.
When the model is fine-tuned, the fingerprints in $S$ and $S'$ are visually indistinguishable.
The reason is that only a few weights of the model are changed by fine-tuning but most weights of the model are not changed.
Thus, the fingerprints of the pirated model have a high similarity with that of the original model.
The experimental results demonstrate that the fingerprints of the model can be successfully extracted and used for ownership verification even though the model is fine-tuned.
In other words, the proposed DNN IP protection method is robust to fine-tuning attack.

\begin{table}[!htbp]
\centering
\caption{Test Accuracy and \textit{ASSIM} after Fine-tuning with Different Epochs.}
\begin{tabular}{ccccc}
\toprule
\textbf{Dataset} & \textbf{DNN model} & \textbf{Epochs} & \textbf{TA} & \textit{ASSIM}\\
\midrule
\multirow{8}[2]*{ImageNet} &
\multirow{4}*{VGG-16}
 &0 & 73.82\% & 1.0000 \\
  && 20 & 80.60\% & 0.9992 \\
  & & 50 & 83.60\% & 0.9989 \\
  & & 80 & 83.70\% & 0.9987 \\
  \cmidrule{2-5}
& \multirow{4}*{ResNet-18}
& 0 & 75.94\% & 1.0000 \\
   & &20 & 78.50\% & 0.9366\\
  & & 50 & 76.93\% & 0.9064\\
  & & 80 & 75.23\% & 0.8964\\
\bottomrule
\end{tabular}%
\label{tab:fine-tune}
\end{table}

\subsubsection{Pruning Attack}\label{prune}
Model pruning \cite{HanPTD15} is used to compress the DNN model by setting $p$ (\%) of the weights with low absolute values to zero.
Since the pruning attack will significantly change the parameters of the model, the extracted fingerprints may be affected.
In this experiment, the robustness of the proposed method against pruning attack is evaluated in terms of test accuracy on clean images and $ASSIM$.
As shown in Table \ref{tab:prune}, under the pruning attack with pruning rate of 0\%, 20\%, 40\% and 60\%, the $ASSIM$ of fingerprints extracted from VGG-16 model and ResNet-18 model is 1.0000, 0.9986, 0.9950, 0.9814 and 1.0000, 0.9902, 0.9669, 0.9322, respectively, which are much higher than the threshold.
This demonstrates that, when the parameters of the model have been pruned with the pruning rate up to 60\%, the extracted fingerprint will not be affected.
As a result, the proposed DNN IP protection method is robust against pruning attack.

The adversary usually will not prune the model with a high pruning rate as the adversary wants to maintain the inference performance of the model.
As shown in Table \ref{tab:prune}, when pruning rate is between 0\%$\sim$40\%, the test accuracy of VGG-16 model and ResNet-18 model (after being pruned) ranges from 73.82\% to 72.82\% and 75.94\% to 72.28\% respectively, which are almost consistent to that of the original model.
However, when the pruning rate is set to 60\%, the test accuracy of VGG-16 model and ResNet-18 model after pruning decrease to 63.04\% and 62.36\%, respectively, which are much lower than that of the original model.
Therefore, the adversary will not prune the model with a high pruning rate (higher than 60\%).

\begin{table}[!htbp]
\centering
\caption{Test Accuracy of Model and \textit{ASSIM} of Fingerprints after Pruning Attack with Different Pruning Rates.}
\begin{tabular}{cccccc}
\toprule
\textbf{Dataset} & \textbf{Model} & \textbf{$p$ (\%)} & \textbf{TA} & \textit{ASSIM}\\
\midrule
\multirow{8}[2]*{ImageNet}&
\multirow{4}*{VGG-16}
  & 0   & 73.82\% & 1.0000\\
 &  & 20 & 73.62\% & 0.9986 \\
 & & 40 & 72.82\% & 0.9950 \\
 & & 60 & 63.04\% & 0.9814 \\
\cmidrule{2-5}
& \multirow{4}*{ResNet-18}
  & 0 & 75.94\% & 1.0000\\
  & & 20 & 75.28\% & 0.9902 \\
  & & 40 & 72.28\% & 0.9669 \\
  & & 60 & 62.36\% & 0.9322 \\
\bottomrule
\end{tabular}%
\label{tab:prune}
\end{table}

\subsubsection{Watermark Overwriting Attack}
In this section, we evaluate the robustness of the fingerprints against watermark overwriting attack \cite{RouhaniCK19}.
The watermark used in this section is the noise-based watermark, which is proposed in \cite{zhang2018protecting}.
The noise-based watermark \cite{zhang2018protecting} is embedded into the VGG-16 and ResNet-18 models respectively.
We use watermark accuracy (WA) \cite{zhang2018protecting} to evaluate the accuracy of the DNN model on the watermark instances, which represents the ratio of the watermark instances classified as the specific class among all the watermark instances.

As shown in Table \ref{tab:woa}, the WA of VGG-16 model and ResNet-18 model is 91.50\% and 97.00\%, respectively, which indicates that the watermark is successfully embedded into the model.
Under the watermark overwriting attack, the $ASSIM$ of fingerprints extracted from VGG-16 model and ResNet-18 model is 0.9991 and 0.9887, respectively.
The reason is that, in order to remain the performance of model, these watermarking methods only modify a few redundant parameters of the model.
Hence, the intrinsic features will not change after the model has been embedded with watermark.
The experimental results demonstrate that the ownership of the model can be successfully verified even though the model is embedded with watermark by the adversary via watermark overwriting attack.
In other words, the proposed DNN IP protection method is robust against watermark overwriting attack.

\begin{table}[!htbp]
\centering
\caption{The Robustness of the Proposed DNN IP Protection Method Against Watermark Overwriting Attack}
\setlength\tabcolsep{3pt}
\begin{tabular}{ccccc}
\toprule
\textbf{Dataset}&\textbf{Model}&\textbf{TA}&\textbf{WA}&\textit{ASSIM}\\
\midrule
\multirow{2}*{ImageNet}
& VGG-16 &  74.10\% & 91.50\% & 0.9991\\
&ResNet-18 & 76.20\% & 97.00\%&  0.9887\\

\bottomrule
\end{tabular}%
\label{tab:woa}
\end{table}

\subsubsection{Adaptive Attack}
\begin{figure}[!htbp]
\centering
\includegraphics[width=0.48\textwidth]{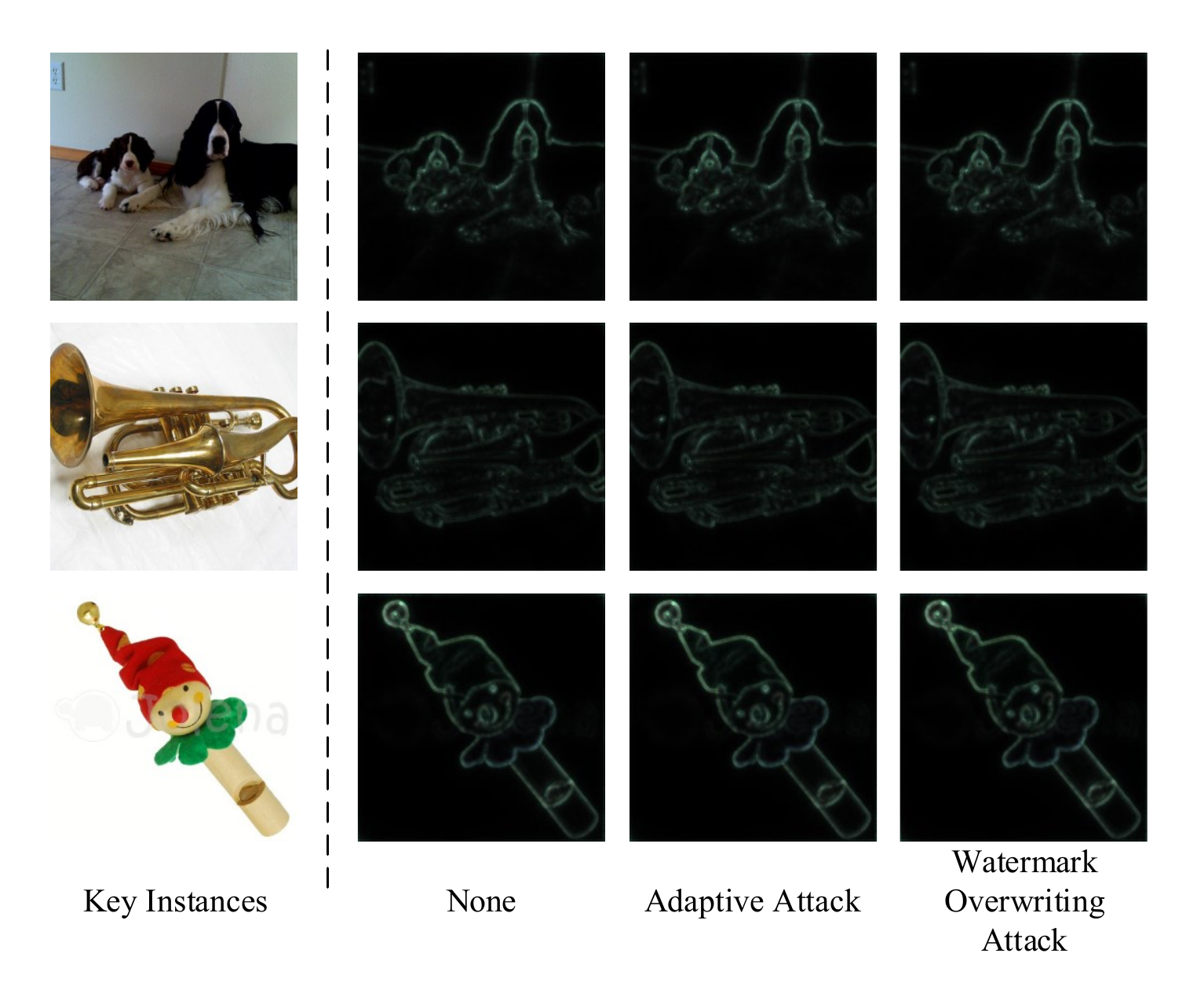}
\caption{Examples of fingerprints generated from VGG-16 model subjected to adaptive attack and watermark overwriting attack.}
\label{fig_robustness}
\end{figure}

In this section, we evaluate the robustness of the proposed DNN IP protection method against adaptive attack.
In this adaptive attack scenario, the adversary is assumed to have knowledge of the approximate protection mechanism.
The adversary knows that the model is verified by extracted fingerprints of the model, but does not know the specific flow of the algorithm.
The adversary may use pruning attack to destroy the watermark and then use fine-tuning to recover the performance of the model.
For the adaptive attack, we first prune the model with 40\% of pruning rate and then fine-tune the model for 50 epochs.
The robustness of the proposed DNN IP protection method against adaptive attack is evaluated on CIFAR-10 and ImageNet datasets respectively.
Fig. \ref{fig_robustness} shows the fingerprints of the protected model and the pirated model, where the pirated model is modified by adaptive attack.
It is shown that, the fingerprint extracted from the model under adaptive attack or watermark overwriting attack is consistent with the fingerprint extracted from the original model.
As shown in Table \ref{tab:adatt}, after adaptive attack, the $ASSIM$ is much greater than the threshold $T$.
Specifically, on CIFAR-10 dataset, when the VGG-16 model and
ResNet-18  model are pruned and fine-tuned, the $ASSIM$ between $S$ and $S'$ is 0.9867 and 0.9637 respectively.
On ImageNet dataset, the $ASSIM$ is 0.9949 and 0.9175 on VGG-16 and ResNet-18 model respectively.
The reason is that most intrinsic features of the model will not change after adaptive attack.
In conclusion, the extracted fingerprints are robust against the adaptive attack.

\begin{table}[!htbp]
\centering
\caption{the Robustness of the Proposed DNN IP Protection Method Against Adaptive Attacks}
\setlength\tabcolsep{3pt}
\begin{tabular}{ccccc}
\toprule
\textbf{Attack}&\textbf{Dataset}&\textbf{Model} &\textbf{TA}&\textit{ASSIM}\\
\midrule
\multirow{4}[3]*{Adaptive Attack}
&\multirow{2}*{CIFAR-10}
&VGG-16& 85.86\% & 0.9867\\
& &ResNet-18& 80.40\% & 0.9637\\
\cmidrule{2-5}
&\multirow{2}*{ImageNet}
& VGG-16 &  80.76\% & 0.9949\\
& &ResNet-18 & 77.30\% & 0.9175\\

\bottomrule
\end{tabular}%
\label{tab:adatt}
\end{table}

\subsection{Comparison with Existing Works}
In this section, we compare our work with existing works.
As shown in Table \ref{discussion}, different DNN IP protection methods are compared in terms of model modification, interpretability, extra training, computational overhead and large-scale datasets.

\begin{table*}[!htbp]
  \centering
  \caption{The Comparison between Existing Works and Our Work}
    \begin{tabular}{cccccc}
    \toprule
    \textbf{Works} & \textbf{Model Modification} & \textbf{Interpretability} & \textbf{Require Extra Training} & \textbf{Computational Overhead} & \textbf{Large-scale Dataset} \\
    \midrule
    Uchida \textit{et al.} \cite{uchida2017embedding} & Yes   & No    & Training from scratch     & High     & No \\
    Rouhani \textit{et al.} \cite{RouhaniCK19} & Yes   & No    & Fine-tuning     & Medium     & No \\
    Adi  \textit{et al.} \cite{adi2018turning} & Yes   & No    & Training from scratch or Fine-tuning     & Medium     & Yes \\
    Zhang \textit{et al.} \cite{zhang2018protecting} & Yes   & No    &  Fine-tuning    & Medium     & No \\
    Xie \textit{et al.} \cite{xie2021deepmark}  & Yes   & No    & Fine-tuning     & Medium     & No \\
    Cao  \textit{et al.} \cite{CaoJG21} & No    & No    & No     & Small     & Yes \\
    Pan \textit{et al.} \cite{PanYZY22}  & No    & No    & Training extra classifier     & High     & Yes \\
    Ours  & No    & Yes   & No     & Small     & Yes \\
    \bottomrule
    \end{tabular}%
  \label{discussion}
\end{table*}%

\textit{Model Modification:}
The work \cite{uchida2017embedding} embeds watermark into the model by using a parameter regularizer.
The work \cite{RouhaniCK19} embeds watermark into the Probability Density Function (PDF) of model by fine-tuning the model.
The work \cite{xie2021deepmark} embeds watermark into the redundant parameters of model by fine-tuning the model.
The works \cite{zhang2018protecting, adi2018turning} inject backdoor into the model as the watermark.
All these works \cite{uchida2017embedding,RouhaniCK19,xie2021deepmark,zhang2018protecting, adi2018turning} require to modify the parameters of the model so as to embed the watermark.
When the length of embedded watermark is large, the test accuracy of watermarked model will degrade compared with the clean model.
As a comparison, the proposed method utilizes intrinsic features of the model as the fingerprints without modifying the model's parameters, which will not affect the inference performance of the model.

\textit{Interpretability:}
Most existing works \cite{uchida2017embedding, RouhaniCK19,zhang2018protecting, adi2018turning, CaoJG21, PanYZY22, xie2021deepmark} lack of interpretability.
In comparison, the proposed method uses intrinsic features of the model for ownership verification.
These intrinsic features are calculated based on the activations of internal neurons.
Therefore, the fingerprints composed of intrinsic features can represent the model in an explainable way.

\textit{Extra Training}:
The work \cite{uchida2017embedding} requires to train the model from scratch so as to embed the model owner's watermark into the model.
The work \cite{RouhaniCK19} fine-tunes the model to embed the watermark.
The works \cite{adi2018turning, zhang2018protecting} utilize backdoor attack to embed watermark into the model through fine-tuning.
The work \cite{xie2021deepmark} locates the redundant neurons through model pruning and then embeds the watermark into these redundant neurons through fine-tuning.
The work \cite{PanYZY22} needs to train an extra model (binary classifier) for ownership verification.
Therefore, all these works \cite{uchida2017embedding, RouhaniCK19, adi2018turning, zhang2018protecting, xie2021deepmark, PanYZY22} require extra training.
In contrast, the proposed method directly extracts the fingerprint from the model without any extra training.

\textit{Computational Overhead}:
The works \cite{uchida2017embedding, RouhaniCK19, adi2018turning, zhang2018protecting, xie2021deepmark} require to train the target model from scratch or fine-tune the target model so as to embed the watermark, which introduce a high or medium computational overhead.
The work \cite{PanYZY22} extracts adversarial examples from target model and other models.
Then, they use these adversarial examples to train an extra model (binary classifier).
Hence, the computational overhead of the work \cite{PanYZY22} is very high.
In comparison, the proposed method extracts the fingerprints from the model without any extra training (which does not require training from scratch or fine-tuning the target model, or even training an extra model).
Hence, the computational overhead of the proposed method is very small and much smaller compared with the works \cite{uchida2017embedding, RouhaniCK19, adi2018turning, zhang2018protecting, xie2021deepmark, PanYZY22}.

\textit{Large-scale Dataset:}
Most existing works \cite{uchida2017embedding, RouhaniCK19, zhang2018protecting, xie2021deepmark} only use small datasets, i.e., MNIST, CIFAR-10, and CIFAR-100, to evaluate the performance of their proposed methods.
As a comparison, the DNN IP protection method proposed in this paper is evaluated on ImageNet dataset, which is a large-scale dataset.

\section{Conclusion}\label{conclusion}
In this paper, the first interpretable DNN intellectual property protection method is proposed by exploiting an explainable artificial intelligence method, which does not require to modify the model.
The extracted intrinsic features are viewed as the fingerprints of the model.
The proposed method is demonstrated to be effective on CIFAR-10 dataset and the real and large dataset, ImageNet.
Meanwhile, the fingerprints are robust against fine-tuning attack, pruning attack, watermark overwriting attack, and adaptive attack.
Compared with existing DNN intellectual property methods, the proposed method significantly reduces the overhead, and provides interpretability for the first time which can meet the urgent need for interpretability in DNN commercial applications.
In the future, we will explore fragile watermark for DNN IP protection.

\bibliographystyle{IEEEtran}
\bibliography{ref}

\end{document}